\pdfoutput=1

\documentclass[11pt]{article}

\usepackage[final]{acl}

\usepackage{times}
\usepackage{latexsym}

\usepackage[T1]{fontenc}

\usepackage[utf8]{inputenc}

\usepackage{microtype}

\usepackage{amsmath}
\usepackage{amssymb}
\usepackage{booktabs}

\title{Can Language Models perform Abductive Commonsense Reasoning?}

\author{Seungone Kim \\
  Yonsei University, Department of Computer Science \\
  \texttt{louisdebroglie@yonsei.ac.kr} \\
}

\begin{document}
\maketitle
\begin{abstract}
Abductive Reasoning is a task of inferring the most plausible hypothesis given a set of observations. In literature, the community has approached to solve this challenge by classifying/generating a likely hypothesis that does not contradict with a past observation and future observation. Some of the most well known benchmarks that tackle this problem are aNLI and aNLG (pronounced as alpha-NLI and alpha-NLG). In this report, I review over some of the methodologies that were attempted to solve this challenge, re-implement the baseline models, and analyze some of the weaknesses that current approaches have. The code and the re-implemented results are available at \href{https://github.com/SeungoneKim/abductive-commonsense-reasoning}{this link}\footnote{https://github.com/SeungoneKim/abductive-commonsense-reasoning}.
\end{abstract}

\section{Introduction}

Reasoning can be divided into three different categories; (1) \textbf{Abduction} is a process of both generating hypotheses and selecting some for further pursuit; (2) \textbf{Deduction} draws out their testable consequences; (3) \textbf{Induction} evaluates them~\citep{kapitan1992peirce}. People unconsciously and constantly perform these three different types of reasoning within daily lives. Reasoning allows us to be intelligent coupled with abstraction capabilities~\citep{chollet2019measure}. In order to accomplish the grand goal of building \textit{Artificial General Intelligence}(AGI), reasoning is one of the biggest obstacles that must be fully understood, but there is less progress made.

Within the research community, there has been various attempts to build systems that could mimic the reasoning capabilities a human might have. More specifically, researchers have paid a lot of attention to \textit{Commonsense Reasoning}, a field that deals with knowledge and inferences that we encounter in everyday situations. Different line of works tried to build benchmarks that test whether a system posses a particular commonsense reasoning type~\citep{qin2019counterfactual, zhang2020winowhy, bisk2020piqa, sap2019social, zellers2019hellaswag, zhou2019going, talmor2019commonsenseqa}. Others tried to build methodologies that are efficient in solving these tasks~\citep{klein2019attention, wang2020connecting, rajani2019explain, majumder2020like}.

\textbf{ARI} is a large-scale benchmark dataset that specifically tackles abductive commonsense reasoning~\citep{bhagavatula2019abductive}. In this setting, two text inputs are given; (1) a observation that is made in the past; (2) a observation that is made in the future. Given the input, the goal of \textbf{aNLI} task is to classify a more plausible hypothesis that might have occurred between the past - future observations given two candidate hypotheses, while the goal of \textbf{aNLG} is to generate this hypothesis. For example, when a person locked up all the windows and doors before leaving the house and observed that the house was set open with all the furniture broken, it is more likely that a thief might have stole from his/her house instead of a bird flying through the house and making a mess. 

Generating a good hypothesis is considered more challenging than selecting a better hypothesis since it tests the model has to generate a plausible text output based on the commonsense knowledge it has and map with with the given observations. It could be viewed of a more general form of abductive reasoning. Among the two tasks, I focus on the generative task (aNLG), in this report.

\section{Related Works and Methods}

\subsection{Baseline Model}
As a baseline model, \citet{bhagavatula2019abductive} proposed a model that conditionally generates by simply concatenating the past observation and future observation back and forth, and divide it with special tokens  <o1>, </o1>, <o2>, </o2>. The loss is formalized as a negative log-likelihood as in a typical text generation setting.

\begin{equation}\label{firstloss}
    \mbox{${\cal L}$} = - \sum_{i=1}^{N} \log P(w_{i}^{h}|w_{<i}^{h}, w_{1}^{o1} ... w_{m}^{o1}, w_{1}^{o2} ... w_{n}^{o2}) 
\end{equation}

In this report, I re-implemented the inference procedure using this model and compare with the reported scores of \citet{bhagavatula2019abductive}.

\subsection{Using a Commonsense Inference Model for assistance}
\citet{bhagavatula2019abductive} proposed another model that conditions on an additional background knowledge ${\cal K}$. Specifically, the authors used a commonsense knowledge inference model COMET~\citep{bosselut2019comet}. COMET is a transformer-based model that generates a commonsense inference when a narrative text and its corresponding relation (e.g., xWant, xNeed, xIntent) is given. With an additional background knowledge ${\cal K}$, the loss can be formalized as follows.

\begin{equation}\label{secondloss}
\begin{split}
    \mbox{${\cal L}$} = - \sum_{i=1}^{N} \log P(w_{i}^{h}|w_{<i}^{h}, w_{1}^{o1} ... w_{m}^{o1}, \\w_{1}^{o2} ... w_{n}^{o2}, {\cal K}) 
\end{split}
\end{equation}

The additional commonsense knowledge can be provided in 2 different methods. First, the text output of the commonsense knowledge can explicitly be provided as additional input where the text generation model can potentially use it to generate a plausible hypothesis. Second, the COMET embedding output of the commonsense inferences can be given as additional input by concatenating with the token embedding of the observations. \citet{bhagavatula2019abductive} use a setting of appending eighteen additional embeddings (corresponding to nine relations for each observation) and explains that this allows the model to learn each token's representation while attending to the COMET embeddings to effectively integrate background commonsense knowledge into a language model.

In this report, I re-implemented the inference procedure using this model and compare with the reported scores of \citet{bhagavatula2019abductive}.

\subsection{Enhanced Decoding Methods}
The approach of finetuning a Pre-trained Language Model on a particular task has gained success in different subfields within Natural Language Processing. However, supervised approaches still perform considerably worse than humans and are subject to developing superficial strategies such as repeatedly generating the same phrase or memorizing prevalent surface patterns specific in the dataset. A number of research tackles this problem and come up with some advanced form of text decoding methods to alleviate this problem.

\citet{qin2020back} proposed a decoding method, DELOREAN that is suitable for abductive reasoning and counter-factual reasoning. Instead of simply concatenating the past and future observations when they are given as input like Equation~\ref{firstloss}, \ref{secondloss}, DELOREAN generates the hypothesis based only on the past observation. Then, by conditioning on the past observation and the hypothesis candidate, the model generates a plausible future observation. Note that while there is a ground truth future observation, a new future observation is additionally generated. The cross entropy loss between the existing future observation and the newly generated future observation is used to compute the gradients of the hypothesis using backpropagtion algorithm. Lastly, the model samples possible tokens step-by-step using both the logits of the forward pass and the backward pass. Therefore, DELOREAN isa backprop-based decoding algorithm.

More recently, \citet{qin2022cold} proposed a different decoding algorithm, COLD which is designed for abductive reasoning, counter-factual reasoning, and constrained reasoning. COLD treats text generation as sampling from an energy-based distribution~\citep{hinton2002training, lecun2006predicting}, which allows flexibly composing constraints based on the task at hand. It uses Langevin dynamics which is capable of mapping continuous sample into discrete, fluent text. It uses the gradient of the energy function which can specifically be designed based on the constraints that exist in the underlying task on-the-fly without the need of fine-tuning.

In this report, I did not re-implemented this methods, but compare with the reported scores of \citet{qin2020back, qin2022cold}.

\begin{table*}[t!]
\begin{center}\small
{\begin{tabular}{l ccccc}
    \toprule
    & \multicolumn{5}{c}{aNLG}\\
    \cmidrule(lr){2-6}
    Model & Bleu-4 & METEOR & ROUGE-L & Cider & Bert-score \\
    \midrule
    GPT2 (unsupervised)~\citep{bhagavatula2019abductive} & 0.00 & 9.29 & 9.99 & 3.34 & 36.69\\
    GPT2 + DELOREAN (unsupervised)~\citep{qin2020back} & 1.60 & / & 19.06 & 7.88 & 41.74\\
    GPT2 + COLD (unsupervised)~\citep{qin2022cold} & \textbf{1.79} & / & \textbf{19.50} & \textbf{10.68} & \textbf{42.67}\\
    \midrule
    GPT2 (supervised)~\citep{bhagavatula2019abductive} & 2.23 & 16.71 & 22.83 & \textbf{33.54} & \textbf{48.74}\\
    GPT2 (supervised; \textit{re-implemented}) & 3.06 & 18.61 & 24.49 & 33.42 & 48.70\\
    GPT2 + COMET-txt~\citep{bhagavatula2019abductive} & 2.29 & 16.73 & 22.51 & 31.99 & 48.46\\
    GPT2 + COMET-txt (\textit{re-implemented}) & 3.13 & 18.63 & \textbf{24.50} & 32.66 & 48.50\\
    GPT2 + COMET-emb~\citep{bhagavatula2019abductive} & 3.03 & 17.66 & 22.93 & 32.00 & 48.52\\
    GPT2 + COMET-emb (\textit{re-implemented}) & \textbf{4.13} & \textbf{19.87} & 23.81 & 30.86 & 48.32\\
    \bottomrule
\end{tabular}}
\end{center}
\caption{Automatic evaluation on aNLG benchmark in terms of different metrics}
\label{automatic_evaluation}
\end{table*}

\section{Details of re-Implementation}
I have used \href{https://github.com/allenai/abductive-commonsense-reasoning}{the implementation}\footnote{https://github.com/allenai/abductive-commonsense-reasoning} provided by AllenAI to reproduce the results and test the fine-tuned models.

I have used GPT-2 XL, a 1.5B parameter model as a baseline model for the re-implementation. The inference procedure uses beam-search based decoding with temperature of 1.0, top\_p value of 0.9. For COMET-text and COMET-emb based models, the COMET relation \textsc{oEffect}, \textsc{oReact}, \textsc{oWant}, \textsc{xAttr}, \textsc{xEffect}, \textsc{xIntent}, \textsc{xNeed}, \textsc{xReact}, \textsc{xWant} were used. More details of the fine-tuning procedure is presented in \citet{bhagavatula2019abductive}.

\section{Evaluation of re-Implementation}

The results of the re-implemented versions and the scores reported in each corresponding papers are shown in Table~\ref{automatic_evaluation}. Since it is not fair to directly compare supervised approaches and unsupervised approaches, we should compare independently.

Using Enhanced decoding approaches show a significant amount of better performance in ROUGE-L and Cider metric, gaining more than twice better performance. However, its performance is still not comparable with fine-tuning methods implying that improving decoding strategies itself could not be a substitute fine-tuning yet. This result addresses open-questions to whether a prominent strategy could be used in abductive reasoning instead of the current fine-tuning approaches in future works.

Comparing the re-implemented version and the reported scores of the supervised methods, the re-implemented versions achieved better scores in 11 out of 15. This proves that my re-implementation of inferencing on fine-tuned methods was successful. Also, comparing the scores between different approaches, the methods that reached the highest scores among different metrics differs implying that using COMET text or embedding isn't very effective compared to the baseline model. This also opens room of future research of how we could inject additional knowledge for abductive reasoning.

\section{Analysis on Failure Cases}\label{failure}
In this section, we look at the failure cases each unsupervised and supervised methods made in solving the aNLG task. Some of the failure cases are presented in Table~\ref{failure}.

\begin{table*}[t!]
\begin{center}\small
{\begin{tabular}{l p{0.1\textwidth} p{0.3\textwidth}}
    \toprule
    \multicolumn{2}{l}{\textbf{Past Observation}} &  \textbf{Future Observation} \\
    \midrule
    I really love to play video games. & & I couldn't play video games until I bought a new one.\\
    \midrule
    \multicolumn{2}{l}{\textbf{Answer Hypothesis}} \\ 
    \multicolumn{2}{l}{Sadly my playstation broke after a flood.} \\
    \multicolumn{2}{l}{\textbf{GPT2 (supervised)}} \\
    \multicolumn{2}{l}{I had to quit playing my favorite video games after a week.} \\
    \multicolumn{2}{l}{\textbf{GPT2 + COMET-txt}} \\
    \multicolumn{2}{l}{I had to order new games from my local game store.} \\
    \multicolumn{2}{l}{\textbf{GPT2 + COMET-emb}} \\ 
    \multicolumn{2}{l}{I had a cracked computer.} \\
    \midrule
    
    Amy had a roommate named Sue. & & They didn't speak to each other for weeks.\\
    \midrule
    \multicolumn{2}{l}{\textbf{Answer Hypothesis}} \\ 
    \multicolumn{2}{l}{Sue was very messy.} \\
    \multicolumn{2}{l}{\textbf{GPT2 (supervised)}} \\
    \multicolumn{2}{l}{Amy asked Sue to come to class on time.} \\
    \multicolumn{2}{l}{\textbf{GPT2 + COMET-txt}} \\
    \multicolumn{2}{l}{Amy asked Sue to come to work on time, but she arrived late.} \\
    \multicolumn{2}{l}{\textbf{GPT2 + COMET-emb}} \\ 
    \multicolumn{2}{l}{Sue had to go to work on time.} \\
    \midrule

    Sam always wanted to save up and buy a computer. & & Sam was devastated when he came home to an empty box.\\
    \midrule
    \multicolumn{2}{l}{\textbf{Answer Hypothesis}} \\ 
    \multicolumn{2}{l}{The computer was delivered on the front porch.} \\
    \multicolumn{2}{l}{\textbf{GPT2 (supervised)}} \\
    \multicolumn{2}{l}{Sam didn't save enough money for the computer.} \\
    \multicolumn{2}{l}{\textbf{GPT2 + COMET-txt}} \\
    \multicolumn{2}{l}{Sam didn't save enough money.} \\
    \multicolumn{2}{l}{\textbf{GPT2 + COMET-emb}} \\ 
    \multicolumn{2}{l}{Sam didn't save up enough money to but one.} \\
    \midrule 
    
    In Fort Worth, we have an event call hit the bricks. & & However, thanks to my training the run was easy.\\
    \midrule
    \multicolumn{2}{l}{\textbf{Answer Hypothesis}} \\ 
    \multicolumn{2}{l}{We had to run several miles.} \\
    \multicolumn{2}{l}{\textbf{GPT2 (supervised)}} \\
    \multicolumn{2}{l}{I trained myself for the run.} \\
    \multicolumn{2}{l}{\textbf{GPT2 + COMET-txt}} \\
    \multicolumn{2}{l}{I trained myself to run very fast.} \\
    \multicolumn{2}{l}{\textbf{GPT2 + COMET-emb}} \\ 
    \multicolumn{2}{l}{I took a deep learning class to teach myself how to run.} \\
    \midrule 
    
    Nikki wanted candy. & & Nikki was very mad at her mother.\\
    \midrule
    \multicolumn{2}{l}{\textbf{Answer Hypothesis}} \\ 
    \multicolumn{2}{l}{Her mother wouldn't buy candy because it's bad for Nikki's teeth.} \\
    \multicolumn{2}{l}{\textbf{GPT2 (supervised)}} \\
    \multicolumn{2}{l}{Nikki got candy from her mom.} \\
    \multicolumn{2}{l}{\textbf{GPT2 + COMET-txt}} \\
    \multicolumn{2}{l}{Nikki got out of the deli.} \\
    \multicolumn{2}{l}{\textbf{GPT2 + COMET-emb}} \\ 
    \multicolumn{2}{l}{She went to the store for some candy.} \\
    \midrule
    
    The cat sunned itself where the light passed through the window. & & The cat slept there fore the next two hours.\\
    \midrule
    \multicolumn{2}{l}{\textbf{Answer Hypothesis}} \\ 
    \multicolumn{2}{l}{The cat made it self comfortable in the sunlight.} \\
    \multicolumn{2}{l}{\textbf{GPT2 (supervised)}} \\
    \multicolumn{2}{l}{The cat was sleeping in a shady spot.} \\
    \multicolumn{2}{l}{\textbf{GPT2 + COMET-txt}} \\
    \multicolumn{2}{l}{The cat was sleeping in a hard pile.} \\
    \multicolumn{2}{l}{\textbf{GPT2 + COMET-emb}} \\ 
    \multicolumn{2}{l}{The cat always woke up in the dark.} \\
    
    \bottomrule
\end{tabular}}
\end{center}
\caption{Failure examples of generated hypotheses from GPT2 and its variants}
\label{failure}
\end{table*}

\subsection{Language Models do not capture the \textit{Causal Relationship} between events}

Majority cases of Language Models failing to capture the \textit{Causal Relationship} between events were found. For example, given a past observation such that 'Sam always wanted to save up and buy a computer' and a future observation such that 'Sam was devastated when he came home to an empty box', we could easily suspect that Sam saved money, bought a computer, but he was scammed and got an empty box. The reasoning chain requires 4 hops in total which is very challenging for a Language Model to learn. Therefore, GPT2, GPT2 + COMET-txt, GPT2 + COMET-emb all generated a hypothesis that 'Sam didn't save enough money'.

In order to overcome this issue, it is likely that more advanced learning methods that could incorporate causal reasoning should be developed. Simply showing the answer hypothesis is not enough to enforce the model to learn the mapping patterns because it is well known that Deep Learning methods are likely to learn the shortcuts~\citep{geirhos2020shortcut}. Although increasing the model size and data size is the top-trend within the community of Natural Language Processing~\citep{brown2020language}, there is room of improvement that could be accomplished by establishing reasoning based learning methods~\citep{luo2020causal}. Simply adding the text outputs or embeddings of commonsense inference models did not resolve the issue of enforcing the model to capture the causal relationships.

\subsection{Language Models are weak in \textit{Negation Logic}}

Negation is an important property in many language understanding tasks, such as sentiment analysis, question answering, knowledge base completion and natural language inference~\citep{hosseini2021understanding}. However, a number of examples generated by GPT2 did not seem to capture the crucial negation logic between the past and future observations. For example, given a past observation such that 'Nikki wanted candy' and a future observation such that 'Nikki was very mad at her mother', we could conjecture that Nikki did not get candy from her mother. However, all of the generated results included the information that Nikki got candy from her mother.

I conjecture there are two possible reasons why GPT2 did not successfully generate an plausible hypothesis. First, it did not understand that in order for Nikki to be angry at her mother, she wouldn't have got what she wanted. This means that the model was unable to capture the negation logic in a social situation. Second, the model might not have the commonsense knowledge to inference the hypothesis. If the model possessed  commonsense knowledge that mothers typically do not give candy to their children, it is more likely to have generated a more acceptable result.

\subsection{Language Models generate open-bounded results in \textit{Open-Domain Tasks}}

The ability to control the generation results is crucial in open-domain settings. There were various methods developed to control the desired output in response generation~\citep{yangprogressive} and text-generation~\citep{keskar2019ctrl,dathathri2019plug}. In abductive commonsense reasoning, the scope of possible outputs is very large, and I observed a lot of cases where the model generates a awkward output. For example, given a past observation such that 'Amy had a roommate named Sue' and a future observation such that 'They didn't speak to each other for weeks' it is likely that some part of Sue upset Amy or vice versa. However, the result in which GPT2 generated is that Amy or Sue did not come to work in time. Although it is possible that Amy and Sue might also be coworkers, a more appropriate answer should be constrained into an event that might have happened inside their room. The models did not  generate a similar output to the answer which is 'Sue was very messy'.

\section{Conclusion}
In this report, I have summarized a novel commonsense reasoning task of abductive reasoning. I explained some of the state-of-the-art methodologies in solving the task and re-implemented the supervised methods and measured their scores in terms of automatic evaluation metrics. I have analyzed some of the weaknesses by introducing some failure cases and suggested how the community could develop more advanced learning methods to make models that could perform abductive reasoning. I have included the code I have used for re-implementation in the report.

\section*{Acknowledgements}

This report is submitted to the final project of BigData course(CSI4121.01) of Yonsei University. All the experiment code and report was written by the author.

\bibliography{anthology,custom}
\bibliographystyle{acl_natbib}

\end{document}